\documentclass[letterpaper, 10 pt, conference]{ieeeconf}  

\IEEEoverridecommandlockouts
\usepackage[T1]{fontenc}
\usepackage{graphicx}
\usepackage{amsmath,amssymb}
\usepackage[dvipsnames]{xcolor}
\usepackage[
  breaklinks=true,
  colorlinks=true,
  linkcolor=MidnightBlue,
  citecolor=MidnightBlue,
  urlcolor=MidnightBlue
]{hyperref}
\usepackage{booktabs}
\usepackage{makecell}
\usepackage{soul}
\usepackage{multirow}
\usepackage{xspace}
\usepackage{tabularx}
\usepackage{colortbl}
\usepackage{cleveref}
\usepackage{siunitx} 
\usepackage[font=footnotesize]{caption}
\usepackage{subcaption}
\usepackage{newtxtext,newtxmath}
\usepackage{pifont}

\newcommand{\cmark}{\ding{51}}
\newcommand{\xmark}{\ding{55}}
\newcommand{\eg}{\emph{e.g.}\xspace}
\newcommand{\ie}{\emph{i.e.}\xspace}

\newcommand{\cf}{cf.\xspace}
\newcommand{\myparagraphnohspace}[1]{\vspace{0.25em}\noindent\textbf{#1}}

\definecolor{highlight}{RGB}{230, 230, 255}
\newcommand{\hcolor}{{\sethlcolor{highlight}\hl{Blue}}}

\newcolumntype{C}{>{\centering\arraybackslash}X}
\newrobustcmd\B{\DeclareFontSeriesDefault[rm]{bf}{b}\bfseries}

\crefname{figure}{Fig.}{Figs.}
\Crefname{figure}{Fig.}{Figs.}
\crefname{table}{Tab.}{Tabs.}
\Crefname{table}{Tab.}{Tabs.}
\crefname{section}{Sec.}{Secs.}
\Crefname{section}{Sec.}{Secs.}
\crefname{subsection}{Sec.}{Secs.}
\Crefname{subsection}{Sec.}{Secs.}
\crefname{equation}{Eq.}{Eqs.}
\Crefname{equation}{Eq.}{Eqs.}

\makeatletter
\let\NAT@parse\undefined
\makeatother
\usepackage[numbers,sort&compress]{natbib}

\title{\LARGE \bf
Multimodal Knowledge Distillation for Egocentric Action Recognition\\Robust to Missing Modalities
}

\author{Maria Santos-Villafranca$^{*1}$ \quad Dustin Carrión-Ojeda$^{*2,3}$ \quad Alejandro Perez-Yus$^{1}$\\
Jesus Bermudez-Cameo$^{1}$ \quad Jose J. Guerrero$^{1}$ \quad Simone Schaub-Meyer$^{2,3}$\\
{\small $^{1}$I3A -- University of Zaragoza \quad $^{2}$Technical University of Darmstadt, Department of Computer Science \quad $^{3}$hessian.AI}\\
{\small $^*$Equal contribution}
}

\begin{document}

\maketitle
\pagestyle{plain}

\begin{abstract}
Egocentric action recognition enables robots to facilitate human-robot interactions and monitor task progress. 
Existing methods often rely solely on RGB videos, although additional modalities, such as audio, can improve accuracy under challenging conditions. 
However, most multimodal approaches assume that all modalities are available at inference time, leading to significant accuracy drops, or even failure, when inputs are missing. 
To address this limitation, we introduce KARMMA, a multimodal \underline{K}nowledge distillation framework for egocentric \underline{A}ction \underline{R}ecognition robust to \underline{M}issing \underline{M}od\underline{A}lities that does not require modality alignment across all samples during training or inference. 
KARMMA distills knowledge from a multimodal teacher into a multimodal student that leverages all available modalities while remaining robust to missing ones, enabling deployment across diverse sensor configurations without retraining. 
Our student uses approximately 50\,\% fewer computational resources than the teacher, resulting in a lightweight and fast model that is well suited for on-robot deployment. 
Experiments on Epic-Kitchens and Something-Something demonstrate that our student achieves competitive accuracy while significantly reducing performance degradation under missing modality conditions. Project page available at: \url{https://visinf.github.io/KARMMA/}
\end{abstract}    
\section{Introduction} \label{sec:introduction}
Egocentric vision aims to capture and interpret the world from a first-person perspective. 
Recently, interest in this field has grown due to advances in human-like robots and the increasing popularity of portable cameras. 
Egocentric vision supports a wide range of applications, including assistive devices~\cite{lee2020hand}, human–robot interaction (HRI)~\cite{fang2024egopat3dv2}, and surveillance~\cite{ardeshir2018integrating}.
The release of large-scale egocentric datasets~\cite{grauman2024ego,damen2018scaling} has further enabled progress in tasks such as gesture/action recognition~\cite{radevski2023multimodal, papanagiotou2021egocentric} and object recognition~\cite{akiva2023self}.

Compared to the exocentric (third-person) setting, egocentric video is more challenging due to camera motion, which introduces blur and frequent occlusions. 
To mitigate these issues, recent works have explored multimodal cues~\cite{radford2021learning,gong2023mmg}.
However, most multimodal methods assume that all modalities are available at inference time. 
This assumption often fails in practice, particularly in robotics, due to privacy constraints, muted microphones, or sensor malfunctions~\cite{wu2024deep} (\eg, a faulty camera). 
Moreover, multimodal architectures typically suffer significant accuracy drops when the most informative modality is missing~\cite{ma2022multimodal}.

\begin{figure}[t]
  \centering 
  \includegraphics[width=\linewidth]{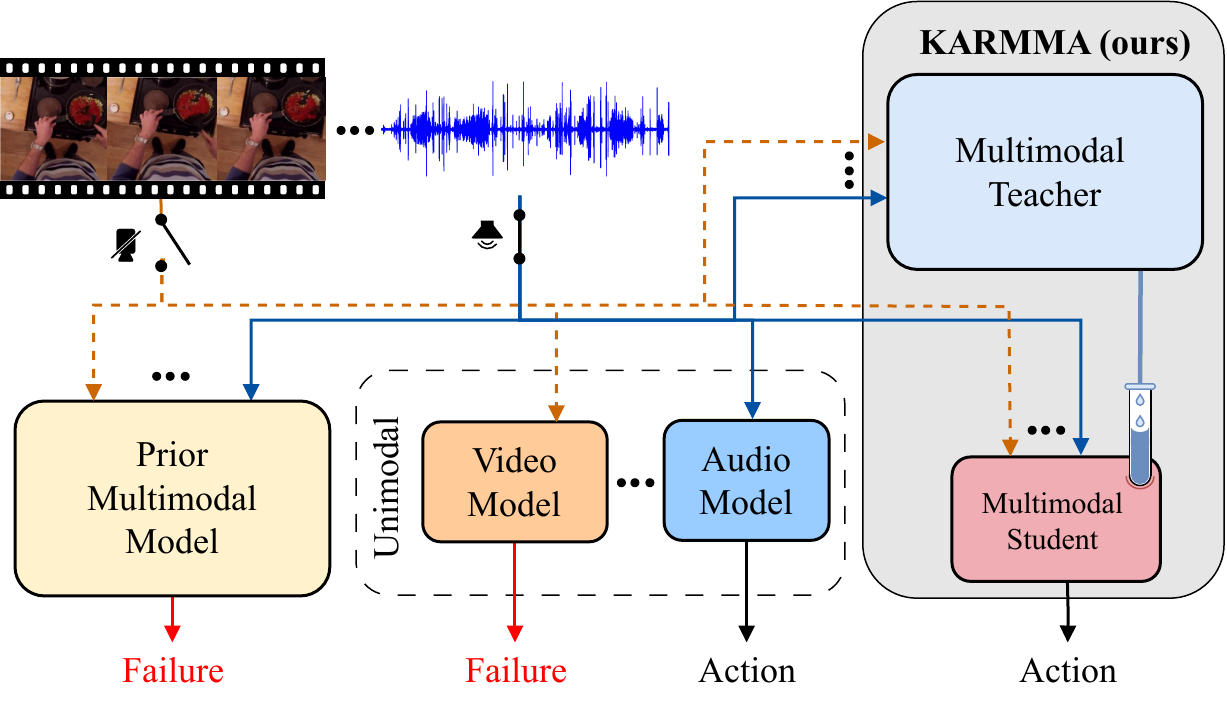}
  \vspace{-1.4em}
  \caption{\textbf{KARMMA motivation.}
  We introduce a novel \emph{multimodal-to-multimodal} framework that leverages all available modalities while remaining robust when any are absent, eliminating the need for modality-aligned data. 
  KARMMA produces a lightweight student that operates on \emph{any} subset of the trained modalities, providing high flexibility and computational efficiency, making it suitable for edge and on-device deployment. 
  Solid lines indicate available modalities, while dashed lines denote missing ones.}
  \label{fig:teaser}
\end{figure}

To meet the safety requirements of robotics, especially HRI, perception must remain reliable under missing modality conditions. 
We address this challenge with a multimodal framework that leverages all available modalities while improving robustness to missing ones.
However, conventional multimodal models are computationally demanding, and processing all modalities can be slow, limiting reliable operation in dynamic or uncertain environments. 
Instead, we propose a multimodal-to-multimodal distillation pipeline in which a large teacher, constructed by fusing frozen, pre-trained unimodal encoders, transfers its knowledge to a lightweight student. 
Notably, both models can operate on any subset of modalities during training and inference, making them suitable for real-world applications where sensor availability is unpredictable (see~\cref{fig:teaser}). 

\myparagraphnohspace{Contributions.} 
\emph{(1)} We propose a novel multimodal-to-multimodal distillation framework for egocentric action recognition that does not require modality alignment across samples during training or inference. 
\emph{(2)} Our distillation process explicitly accounts for missing modalities, producing a lightweight, fast, and flexible student suitable for on-robot deployment.
\emph{(3)} Our teacher fuses features from frozen, pre-trained unimodal encoders, eliminating the need to retrain them, which simplifies the integration of newer encoders as they become available.
\emph{(4)} We introduce a fusion block with a simple, parameter-free token reduction strategy that lowers computational cost without sacrificing accuracy. 

\section{Related Work} \label{sec:related-work}
\myparagraphnohspace{Egocentric action recognition} has primarily been studied using RGB video~\cite{wang2023ego}, which is the most informative modality for this task. 
However, several works show that incorporating additional modalities, such as gaze~\cite{min2021integrating} or audio~\cite{kazakos2021little}, improves accuracy.
Early multimodal research focused on bimodal setups, but the recent release of multimodal egocentric datasets~\cite{grauman2024ego,goyal2017something,damen2018scaling,grauman2022ego4d} has driven the development of multimodal approaches beyond two modalities.
For instance, Gong et al. \cite{gong2023mmg} investigate generalization to unseen and missing modalities, while Dong et al. \cite{dong2023simmmdg} study domain generalization under missing modality conditions. 
Additionally, several works~\cite{radevski2023multimodal,hatano2024multimodal} distill a multimodal teacher into a unimodal student to improve its accuracy. 
In contrast, we distill knowledge from a large multimodal teacher into a compact multimodal student that is explicitly designed to handle missing modalities.

\myparagraphnohspace{Modality fusion} is commonly implemented using cross-attention~\cite{pramanick2023egovlpv2}, although such pairwise fusion limits scalability beyond two modalities. 
Contrastive objectives provide an alternative strategy for fusing modality pairs. 
For example, Lin et al. \cite{lin2022egocentric} introduce a contrastive loss tailored to egocentric video-language alignment. 
Other approaches pre-train modality-specific experts and fuse their frozen features using model-agnostic mechanisms~\cite{radevski2023multimodal}, enabling fusion across multiple modalities. 
However, all these methods assume that all modalities are available at inference time. 
To address this limitation, Gong et al.~\cite{gong2023mmg} apply modality dropout during training to improve robustness to missing inputs, while Ramazanova et al.~\cite{ramazanova2024exploring} introduce a learnable modality token activated only when a modality is absent. 
Ramazanova et al.~\cite{ramazanova2025testtime} further propose a test-time adaptation strategy for missing modalities which requires a separate forward pass for each modality combination and is restricted to dropping a fixed modality in a bimodal setting.
Our transformer‐based fusion extends these ideas by combining modality dropout with two types of learnable tokens. 
The resulting student model is fast and handles any subset of available modalities without requiring additional forward passes.

\myparagraphnohspace{Token reduction} is essential in self-attention-based fusion methods, as computational cost scales quadratically with the number of input tokens.
To mitigate this cost, Fayyaz et al.~\cite{fayyaz2022adaptive} propose a parameter-free module that prunes the attention matrix, while Shang et al.~\cite{shang2024llava} prune tokens via clustering using an interquartile range scoring function~\cite{boukerche2020outlier}. 
Other approaches merge redundant tokens based on cosine similarity~\cite{bolya2022token} or treat tokens as samples of a continuous signal and subsample them accordingly~\cite{marin2023token}. 
In contrast, we show that a simple parameter‐free strategy, averaging contiguous tokens within each modality, effectively reduces computation without sacrificing accuracy.

\myparagraphnohspace{Knowledge distillation} transfers knowledge from a large teacher model to a smaller student. 
In multimodal settings, Radevski et al.~\cite{radevski2023multimodal} and Hatano et al.~\cite{hatano2024multimodal} improve unimodal students using multimodal teachers, while Wei et al.~\cite{wei2023mmanet} propose a multimodal-to-multimodal distillation framework with modality dropout for classification and segmentation. 
Building on these works, we introduce a novel multimodal‐to‐multimodal distillation framework for egocentric action recognition that combines modality dropout with an additional mechanism to enhance robustness to missing inputs.

\section{KARMMA} \label{sec:karmma}
\begin{figure*}[!t]
    \centering
    \includegraphics[width=\linewidth]{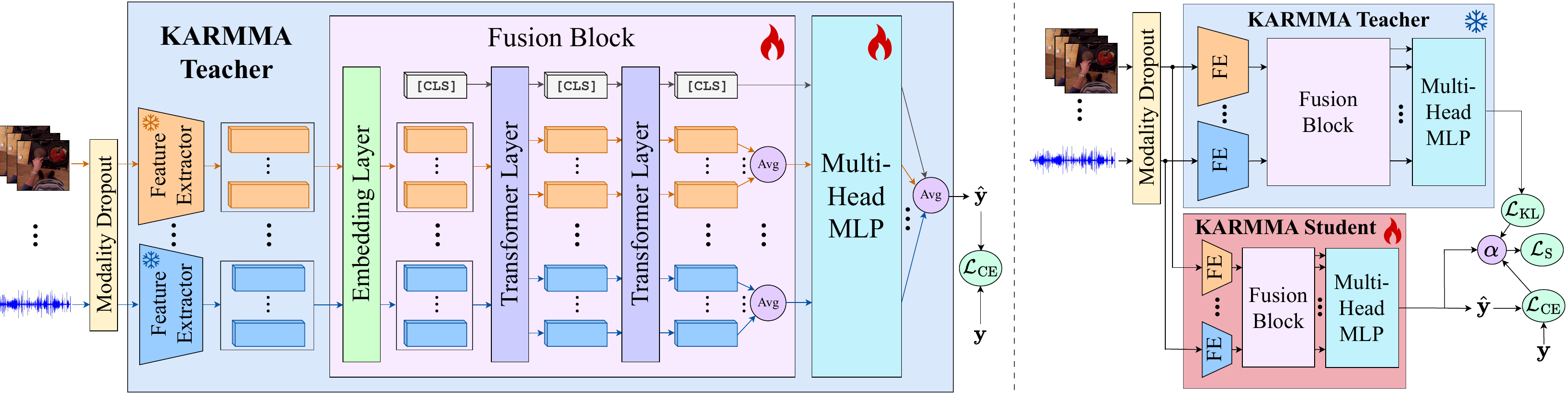}
    \vspace{-1.2em}
    \caption{\textbf{KARMMA training.} 
    \emph{Left} (first stage): training the teacher. 
    \emph{Right} (second stage): distilling knowledge from the frozen teacher into the student. 
    Both networks use modality dropout and our student includes our strategy for handling missing modalities (see~\cref{sec:karmma-enhancements}) to remain robust when inputs are incomplete.}
    \label{fig:karmma-framework}
\end{figure*}

This work addresses multimodal egocentric action recognition, which aims to classify human actions from multiple egocentric modalities. 
\cref{fig:karmma-framework} summarizes our proposed KARMMA framework, which distills knowledge from a large teacher into a compact student.

\subsection{Problem Definition} \label{sec:problem-definition}
Given an egocentric dataset with $M$ modalities, let $\mathbf{x}^j_i \in \mathbb{R}^{T \times D^j_1 \times \cdots \times D^j_n}$ denote the $i^\text{th}$ action sequence of the $j^\text{th}$ modality, where $T$ is the number of time steps and $D^j_1 \times  \cdots \times D^j_n$ are the $n$ modality-specific spatial or spectral dimensions. 
For a multimodal action sequence $\mathbf{X}_i = \left\{ \mathbf{x}^1_i, \mathbf{x}^2_i, \dots, \mathbf{x}^M_i \right\}$, the goal is to learn a model $f$ that predicts the corresponding action $\mathbf{y}_i$, formulated as $\mathbf{\hat{y}}_i = \operatorname{argmax} \, \sigma \left(f \left (\mathbf{X}_i \right)\right)$, where $\sigma(\cdot)$ denotes the softmax operator. 
Since $\mathbf{X}_i$ may contain any subset of the $M$ modalities, $f$ must be robust to missing inputs. 
The representation of actions $\mathbf{y}_i$ also varies across datasets. 
For example, in Something-Something~\cite{goyal2017something}, an action is represented by a single label, whereas in Epic-Kitchens~\cite{damen2018scaling}, it consists of two outputs (noun and verb).

\subsection{Teacher and Student} \label{sec:teacher-and-student}
Both teacher and student consist of three components: modality‐specific feature extractors (FEs), a transformer-based fusion block (FB), and a multi-head MLP (MH-MLP). 
A key advantage is that both models can perform inference on \emph{any} subset of the training modalities, so a single model handles all modality combinations. 
Additionally, the student uses smaller FEs and a more compact FB, reducing memory usage and inference time.  

\myparagraphnohspace{Feature Extractors (FEs).} 
Since $\mathbf{X}_i$ may include up to $M$ modalities, we use $M$ \emph{frozen}, pre-trained \emph{unimodal} FEs for the teacher and train only the remaining components. 
This design facilitates replacing FEs as newer encoders become available. 
For each modality, we select an established FE available in multiple sizes so the teacher uses a larger variant, while the student uses a smaller one. 
When available, student FEs are initialized from pre-trained weights and fine-tuned.

\myparagraphnohspace{Fusion Block (FB).} 
Inspired by previous works~\cite{ma2022multimodal,nagrani2021attention,gong2023mmg}, we design a transformer-based FB capable of handling an arbitrary number of input tokens and modalities. 
As shown in~\cref{fig:karmma-framework} (left), modality-specific tokens are first projected to a shared dimension via a linear embedding layer. 
The projected tokens, together with a learnable \verb|[CLS]| token, pass through $l$ transformer layers. 
The FB outputs $M+1$ tokens: one averaged token per modality and the \verb|[CLS]| token that aggregates cross-modal information, providing a modality-agnostic representation that remains informative when some modalities are missing.

\myparagraphnohspace{Multi-Head MLP (MH-MLP).} 
The $M+1$ tokens produced by the FB are fed into a MH-MLP consisting of a shared fully connected layer followed by independent classification heads. The number of heads matches the number of outputs required to describe an action. 
As discussed in~\cref{sec:problem-definition}, actions may require a single output (\eg, Something-Something~\cite{goyal2017something}) or multiple outputs (\eg, Epic-Kitchens~\cite{damen2018scaling}). 
For each head, we average the $M+1$ predictions to obtain the final output.

\subsection{Token Reduction Strategy} \label{sec:token-reduction-strategy}
Since the computational cost of our FB grows with the number of modalities, we introduce a simple \emph{parameter-free} token reduction strategy ($\Theta$-Average) that caps the number of tokens per modality with a threshold $\Theta$. 
If a modality-specific FE outputs $k$ tokens with $k > \Theta$, we partition them into $\Theta$ groups of $\left\lfloor \frac{k}{\Theta} \right\rfloor$ tokens, assigning the remaining $k \bmod \Theta$ tokens to the last group. 
Averaging tokens within each group yields exactly $\Theta$ tokens. 
If $k \le \Theta$, all tokens remain unchanged. 
Applying our $\Theta$-Average to the outputs of both teacher and student FEs limits the number of tokens entering the FB, reducing computational and memory costs without introducing learnable parameters.

\subsection{KARMMA Enhancements} \label{sec:karmma-enhancements}
Our proposed KARMMA framework incorporates three main enhancements: 
\emph{(1)} modality dropout applied to both teacher and student, 
\emph{(2)} a novel student-side strategy for handling missing modalities, and 
\emph{(3)} a multimodal-to-multimodal distillation scheme that improves the accuracy of a flexible, lightweight, and fast student, eliminating the need for using a large, slow model at inference.

\label{sec:modality-dropout} \myparagraphnohspace{Modality Dropout.}
Previous works~\cite{wei2023mmanet,gong2023mmg,ramazanova2024exploring} show that  modality dropout improves robustness to missing modalities.
Similar to neuron dropout, modality dropout removes entire modalities with probability $p$ while ensuring that at least one modality remains active. 
As illustrated in~\cref{fig:karmma-framework} (right), we apply modality dropout to \emph{both} teacher and student so that neither relies on the full modality set during training. 
This allows KARMMA to train on datasets where action sequences $\mathbf{X}_i$ may contain different modality subsets (\ie, no modality alignment is required), making our framework highly flexible and suitable for practical scenarios, including robotics, where sensor dropouts are common.

\begin{figure}[t]
    \centering
    \includegraphics[width=\linewidth]{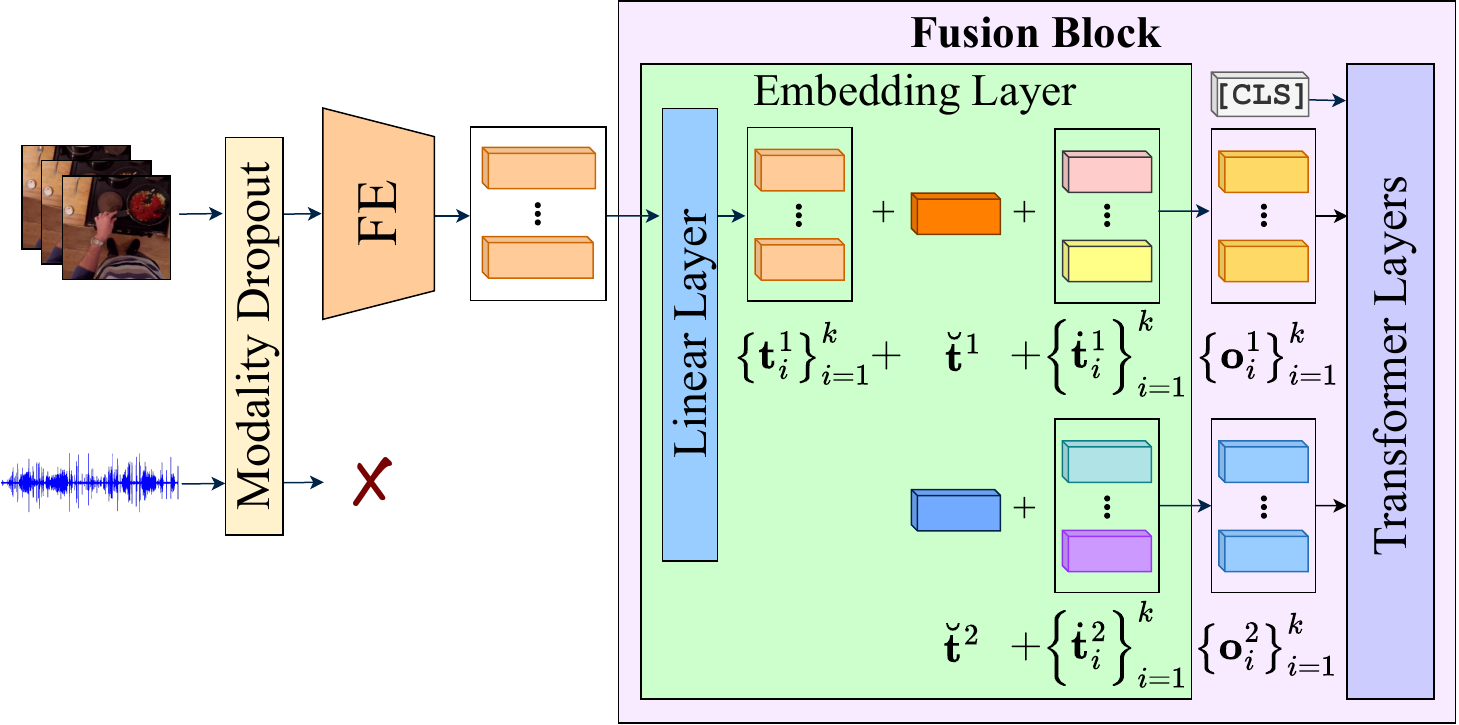}
    \vspace{-1.2em}
    \caption{\textbf{Missing modality strategy.} 
    To handle missing modalities, the embedding layer projects the tokens from the feature extractor when available. 
    Then, it adds a learned modality token $\mathbf{\breve{t}}^m$ to all projected tokens. 
    Finally, a learned token $\mathbf{\dot{t}}^m_i$ is added to each token.}
    \label{fig:missing-modality-strategy}
\end{figure}

\label{sec:missing-modality-strategy} \myparagraphnohspace{Missing Modality Strategy.}
To further enhance the robustness of the student to missing modalities, we propose a simple yet effective strategy illustrated in~\cref{fig:missing-modality-strategy}.
Our strategy introduces two types of learnable tokens:
\begin{enumerate}
    \item \textit{Modality-specific token} $\left( \mathbf{\breve{t}}^m \right)$: one token per modality that helps distinguish modalities and acts similar to positional encodings.
    \item \textit{Token-specific tokens} $\left( \dot{\mathbf{t}}^m_i \right)$: a set of $k$ tokens per modality that compensate when a modality is absent.
\end{enumerate}
Formally, let the FE for modality $m$ produce $k$ tokens from the input $\mathbf{x}^m_i$. Then the output of the embedding layer $\mathbf{o}^m_i$ is expressed as
\begin{equation}
\mathbf{o}^m_i = \begin{cases}
\left\{\mathbf{t}^m_i + \mathbf{\breve{t}}^m + \dot{\mathbf{t}}^m_i \right\}_{i=1}^k & \operatorname{if} \ m \text{ is present},\\
\left\{ \mathbf{\breve{t}}^m + \dot{\mathbf{t}}^m_i \right\}_{i=1}^k & \text{otherwise},
\end{cases}
\label{eq:token-output}
\end{equation}
where $\mathbf{t}^m_i$ is the $i^\text{th}$ projected token for modality $m$.
This strategy provides rich information by combining both types of learnable tokens while keeping the input size of the FB unchanged, allowing the network to effectively handle any pattern of missing modalities.

\label{sec:knowledge-distillation} \myparagraphnohspace{Knowledge Distillation.} 
As shown in~\cref{fig:karmma-framework}, KARMMA operates in two stages. 
First, we train a large multimodal teacher with frozen FEs using the cross-entropy loss:
\begin{equation}
\mathcal{L}_{\text{CE}}= - \frac{1}{\mathit{N}} \sum_{i=1}^{\mathit{N}} \mathbf{y}_i \cdot \log \mathbf{\hat{y}}_i^T,
\label{eq:CE}
\end{equation}
where $\mathit{N}$ is the number of training action sequences, $\mathbf{y}_i$ is the ground-truth action, and $\mathbf{\hat{y}}_i^T$ is the predicted action.

In the second stage, the trained teacher is frozen and its knowledge is distilled to a lightweight student by aligning the class probability distributions from both models via the Kullback–Leibler (KL) divergence:
\begin{equation}
\mathcal{L}_{\textrm{KL}}=\frac{1}{N}\sum_{i=1}^N \mathbf{\bar{y}}^\text{T}_i \cdot \left(\log \mathbf{\bar{y}}^\text{T}_i - \log \mathbf{\bar{y}}^\text{S}_i \right),
\label{eq:KL}
\end{equation}
where $\mathbf{\bar{y}}^\text{T}_i = \sigma\left(f^\text{T} \left ( \mathbf{X}_i \right)\right)$ and $\mathbf{\bar{y}}^\text{S}_i = \sigma\left(f^\text{S} \left ( \mathbf{X}_i \right)\right)$ are the class probability distributions from the teacher and student, respectively. 
To prevent the student from just copying the teacher, we optimize a combination of the cross-entropy and distillation losses: 
\begin{equation}
\mathcal{L}_{\text{S}} = \alpha \, \mathcal{L}_{\textrm{CE}} + (1 - \alpha) \, \mathcal{L}_{\textrm{KL}}\, ,
\label{eq:student-loss}
\end{equation}
where $\alpha$ balances the loss terms. The full student, including its FEs, is trained with this objective, resulting in a model with improved accuracy while requiring substantially less memory and offering faster inference than the teacher.

\section{Experiments} \label{sec:experiments}
\myparagraphnohspace{Datasets.} 
We evaluate our KARMMA framework on two standard egocentric action recognition datasets: Epic-Kitchens-100~\cite{damen2018scaling} and Something-Something V2~\cite{goyal2017something}. 
Epic-Kitchens contains 300 nouns and 97 verbs, with each action defined as a verb-noun pair (\eg, ``pick up + knife''). 
For this dataset, we consider three modalities: RGB video (V), optical flow (F), and audio (A). 
In contrast, Something-Something contains 174 object-agnostic actions (\eg, ``moving \verb|[something]| up'') and provides three modalities: V, F, and object detection annotations (D).

As in most previous works, we report results on the validation splits of both datasets because the official test sets are not publicly available.
However, we observed that the Epic-Kitchens validation split contains four nouns absent from the training set; we therefore remove the corresponding samples and evaluate on this \emph{pruned} validation split. 
Moreover, to avoid overestimating the generalization capabilities of our method by validating and testing on the same data, we further define \emph{Epic-Kitchens\textsuperscript{*}}: a 90\,\%\,/\,10\,\% train/validation split of the original training set, reserving the pruned validation split exclusively for testing.\footnote{The Epic-Kitchens\textsuperscript{*} split is available in the released codebase.}

\myparagraphnohspace{Implementation Details.} 
For the \emph{teacher}, we use Swin-B~\cite{liu2021swin} pre-trained on Kinetics-400~\cite{kay2017kinetics} as the feature extractor for V, A, and F, and a 12-layer STLT~\cite{radevski2021revisiting} pre-trained on Action Genome~\cite{ji2020action} for D. 
The fusion block uses an embedding dimension of 768, 2 transformer layers ($l{=}2$), 8 attention heads, 30\,\% attention dropout, and 50\,\% modality dropout. 
We apply $\Theta$-Average token reduction with $\Theta{=}300$ (see~\cref{sec:token-reduction-strategy}). 
The teacher is trained using AdamW~\cite{loshchilov2019decoupled} for 100 epochs using a batch size of 32, weight decay of 0.05, and gradient clipping at 1.0. 
The learning rate starts at $1e^{-5}$, warms up linearly to $5e^{-4}$ over 10 epochs, and then follows cosine decay back to the initial value.

Unless stated otherwise, the \emph{student} inherits the hyperparameters of the teacher. 
It uses Swin-T~\cite{liu2021swin} pre-trained on Kinetics-400~\cite{kay2017kinetics} for V and F, AST-T~\cite{gong2021ast} for A, and a 9-layer STLT~\cite{radevski2021revisiting} for D.
Since the latter two lack publicly available pre-trained weights, they are trained from scratch. 
The fusion block is reduced to an embedding dimension of 384, with $l{=}1$, and no attention dropout. 
The student is trained with a batch size of 6, a weight decay of 0.01, gradient clipping at 2.0, and a peak learning rate of $1e^{-4}$. 
In~\cref{eq:student-loss}, we set $\alpha{=}0.7$.

\subsection{Analysis of KARMMA} \label{sec:analysis-of-karmma}

\begin{table}[t!]
\centering
\footnotesize
\setlength{\tabcolsep}{5pt}
\caption{\textbf{Analysis of KARMMA under different modality combinations.} 
``KARMMA$_\text{T}$'' and ``KARMMA$_\text{S}$'' denote our teacher and student, respectively (see~\cref{sec:teacher-and-student}).  
``Baseline'' uses the same architecture as ``KARMMA$_\text{S}$'' but is trained end-to-end with cross-entropy loss and without the KARMMA enhancements (see~\cref{sec:karmma-enhancements}), whereas ``Baseline w/ $\delta$'' incorporates modality dropout and our missing modality strategy. 
``V,'' ``F,'' ``A,'' and ``D'' denote RGB video, optical flow, audio, and object detection annotations, respectively, and ``[A/D]'' indicates that either audio or object detection is used. 
We report top-1 action recognition accuracy (\%) on all datasets. 
The ``Epic-Kitchens\textsuperscript{*}'' column reports results on our custom split (see~\cref{sec:experiments}). 
\hcolor{} highlights our final teacher and student. 
\textbf{Bold} and \underline{underlined} values indicate the best and second-best results.}
\label{tab:analysis-of-karmma}
\vspace{-0.2em}
\begin{tabularx}{\columnwidth}{>{\hspace{-\tabcolsep}\raggedright\columncolor{white}[\tabcolsep][\tabcolsep]}lCS[table-format=2.2]S[table-format=2.2]S[table-format=2.2]}
\toprule
\textbf{Method} & {\makecell{\textbf{Inference}\\ \textbf{Modalities}}} & {\makecell{\textbf{Epic-}\\ \textbf{Kitchens}}} & {\makecell{\textbf{Epic-}\\ \textbf{Kitchens\textsuperscript{*}}}} & {\makecell{\textbf{Something-}\\ \textbf{Something}}} \\
\midrule
Baseline & & 40.00 & 39.26 & 57.31 \\
Baseline w/ $\delta$ &  & \underline{41.49} & \underline{40.05} & \underline{60.31} \\
\rowcolor{highlight}
KARMMA$_\text{S}$ &  & \bfseries 43.00 & \bfseries 41.98 & \bfseries 62.88 \\
\rowcolor{highlight}
KARMMA$_\text{T}$ & \smash{\raisebox{1.8em}{V+F+[A/D]}} & 36.44 & 35.84 & 53.51 \\
\midrule
Baseline & & 36.80 & 36.07 & 56.98 \\
Baseline w/ $\delta$ &  & \underline{39.69} & \underline{39.00} & \bfseries 60.38 \\
\rowcolor{highlight}
KARMMA$_\text{S}$ &  & \bfseries 41.29 & \bfseries 40.42 & \underline{58.34} \\
\rowcolor{highlight}
KARMMA$_\text{T}$ & \smash{\raisebox{1.8em}{V+F}} & 32.96 & 31.99 & 43.14 \\
\midrule
Baseline &  & 37.48 & 36.60 & 40.03 \\
Baseline w/ $\delta$ &  & \underline{39.36} & \underline{37.76} & \underline{53.04} \\
\rowcolor{highlight}
KARMMA$_\text{S}$ &  & \bfseries 40.84 & \bfseries 40.10 & \bfseries 59.35 \\
\rowcolor{highlight}
KARMMA$_\text{T}$ & \smash{\raisebox{1.8em}{V+[A/D]}} & 33.79 & 33.44 & 49.89 \\
\midrule
Baseline & & 5.49 & 6.84 & 37.35 \\
Baseline w/ $\delta$ &  & \underline{27.09} & \underline{27.21} & \underline{50.78} \\
\rowcolor{highlight}
KARMMA$_\text{S}$ &  & \bfseries 30.36 & \bfseries 28.24 & \bfseries 53.96 \\
\rowcolor{highlight}
KARMMA$_\text{T}$ & \smash{\raisebox{1.8em}{F+[A/D]}} & 20.44 & 19.95 & 40.62 \\
\midrule
Baseline &  & 32.10 & 32.50 & 39.91 \\
Baseline w/ $\delta$ &  & \underline{37.84} & \underline{36.48} & \bfseries 53.02 \\
\rowcolor{highlight}
KARMMA$_\text{S}$ &  & \bfseries 38.85 & \bfseries 38.37 & \underline{51.97} \\
\rowcolor{highlight}
KARMMA$_\text{T}$ & \smash{\raisebox{1.8em}{V}} & 29.44 & 29.34 & 33.80 \\
\midrule
Baseline & & 2.58 & 4.73 & 36.50 \\
Baseline w/ $\delta$ &  & \underline{25.64} & \underline{26.01} & \bfseries 50.97 \\
\rowcolor{highlight}
KARMMA$_\text{S}$ &  & \bfseries 27.28 & \bfseries 26.06 & \underline{47.24} \\
\rowcolor{highlight}
KARMMA$_\text{T}$ & \smash{\raisebox{1.8em}{F}} & 15.51 & 15.70 & 23.82 \\
\midrule
Baseline &  & 2.05 & 2.25 & 0.07 \\
Baseline w/ $\delta$ &  & \hspace{0.5em} 6.23 & \hspace{0.5em} 6.44 & \hspace{0.5em} 1.21 \\
\rowcolor{highlight}
KARMMA$_\text{S}$ &  & \hspace{0.5em} \underline{8.17} & \hspace{0.5em} \underline{7.53} & \bfseries 37.95 \\
\rowcolor{highlight}
KARMMA$_\text{T}$ & \smash{\raisebox{1.8em}{[A/D]}} & \bfseries 9.38 & \bfseries 9.11 & \underline{29.40} \\
\bottomrule
\end{tabularx}
\end{table}

To evaluate the effectiveness of KARMMA, \cref{tab:analysis-of-karmma} reports the top-1 action recognition accuracy for the teacher (KARMMA$_\text{T}$) and student (KARMMA$_\text{S}$) introduced in~\cref{sec:teacher-and-student}. 
KARMMA$_\text{S}$ includes the enhancements described in~\cref{sec:karmma-enhancements}; thus, to isolate their contributions, we compare against two baselines: 
\emph{(1)}~``Baseline'' uses the same architecture as KARMMA$_\text{S}$ but is trained end-to-end with cross-entropy and without KARMMA enhancements; 
\emph{(2)}~``Baseline w/ $\delta$'' adds modality dropout and our missing modality strategy but excludes knowledge distillation.

The results in~\cref{tab:analysis-of-karmma} show that both baselines and the student outperform the teacher across most modality combinations. 
While knowledge distillation typically produces a stronger teacher than student, in our setting the student benefits from fully trainable feature extractors, whereas the teacher's remain frozen.
Thus, although KARMMA$_\text{S}$ uses smaller feature extractors, they are fine-tuned for multimodal egocentric action recognition. 
In contrast, KARMMA$_\text{T}$ focuses solely on learning effective feature fusion under missing modality conditions. 
Consequently, transferring this knowledge to KARMMA$_\text{S}$ yields absolute gains (relative gains in parentheses) of 3.00\,\% (7.5\,\%), 2.72\,\% (6.9\,\%), and 5.57\,\% (9.7\,\%) over the Baseline when all modalities are present on Epic-Kitchens, Epic-Kitchens\textsuperscript{*}, and Something-Something, respectively. 
Additionally, the results on Epic-Kitchens\textsuperscript{*} closely match those on the original split, indicating robust generalization across data splits.

\cref{tab:analysis-of-karmma} also shows that the highest accuracy is achieved when all modalities are available, confirming that our model effectively leverages multimodal information. 
Moreover, the results confirm that RGB video is the most informative modality, while audio and object detection annotations contribute the least. 
Consequently, replacing V+[A/D] with F+[A/D] leads to a significant accuracy drop across all models, and unimodal [A/D] settings degrade even further. 
However, these drops are more severe for the Baseline model, showing that modality dropout and our missing modality strategy help mitigate over-reliance on dominant modalities.

Since KARMMA$_\text{S}$ outperforms both baselines across most modality combinations and datasets, it demonstrates that while a model can be trained to handle missing modalities (Baseline w/ $\delta$), incorporating our knowledge distillation further improves both accuracy and robustness to missing modalities. 
This improvement is particularly strong on Something-Something, where KARMMA$_\text{S}$ achieves an absolute accuracy gain of 36.74\,\% ($\approx$ 3000\,\% relative) over Baseline w/ $\delta$ when only object detection annotations (D) are used. 
On Epic-Kitchens, the gain is smaller at 1.94\,\% (31\,\% relative), but it still represents a substantial improvement over Baseline w/ $\delta$ when only audio (A) is used. 
The large difference in gains between the two datasets may be due to audio being less informative than object detection annotations, which provide spatial and dimensional information about the objects involved in the action. 

\begin{figure}[t]
  \centering
  \begin{subfigure}{\linewidth}
    \includegraphics[width=1\textwidth]{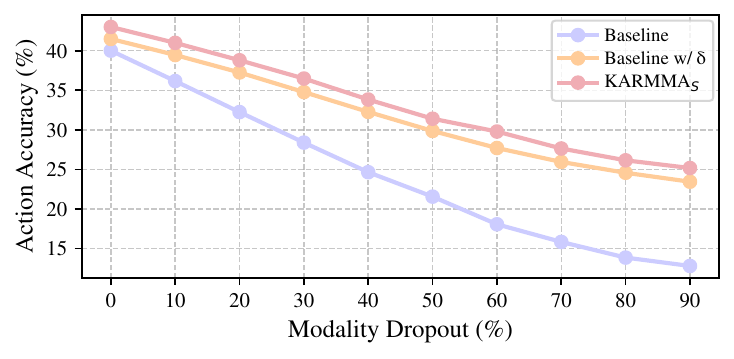}
    \vspace{-1.8em}
    \caption{\footnotesize Epic-Kitchens}
    \vspace{0.1em}
  \end{subfigure}
  \begin{subfigure}{\linewidth}
    \includegraphics[width=1\textwidth]{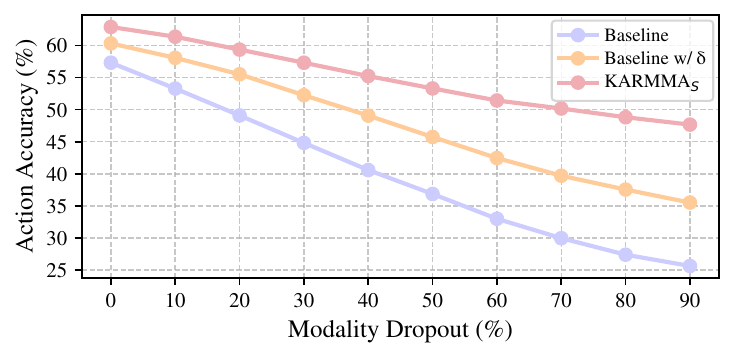}
    \vspace{-1.8em}
    \caption{\footnotesize Something-Something}
  \end{subfigure}
  \vspace{-1.2em}
  \caption{\textbf{Impact of run-time sensor dropouts.} 
  To emulate robotics deployments, we vary the modality dropout probability at inference from 0\,\% to 90\,\% and report action recognition accuracy. 
  ``Baseline'' uses the same architecture as our student ``KARMMA$_\text{S}$'' (see~\cref{sec:teacher-and-student}) but is trained end-to-end with cross-entropy loss and without the KARMMA enhancements (see~\cref{sec:karmma-enhancements}), whereas ``Baseline w/ $\delta$'' incorporates modality dropout and our missing modality strategy.}
  \label{fig:impact-missing-modalities-inference}
\end{figure} 

Although \cref{tab:analysis-of-karmma} demonstrates that our student can effectively handle missing modalities, it only considers scenarios where entire modalities are absent at inference time. 
In contrast, robotics deployments often face run-time \emph{sensor dropouts} (\eg, occluded cameras or muted microphones).
To simulate such conditions, \cref{fig:impact-missing-modalities-inference} reports the action recognition accuracy of the two baselines and our student as the probability of dropping each modality increases from 0\,\% to 90\,\% during inference. 
These results show that the Baseline model is highly sensitive: missing one or more modalities in only 10\% of the samples already reduces accuracy by roughly 5\,\% on both datasets, and increasing this probability to 90\,\% leads to absolute drops of about 27\,\% and 32\,\% for Epic-Kitchens and Something-Something, respectively. 
Incorporating modality dropout and our missing modality strategy enhances robustness, as Baseline w/ $\delta$ exhibits a significantly smaller accuracy drop than the Baseline. 
Moreover, our distillation pipeline consistently improves the accuracy of KARMMA$_\text{S}$, yielding absolute gains (relative gains in parentheses) of approximately 2\,\% (9\,\%) and 12\,\% (34\,\%) over Baseline w/ $\delta$ in the 90\% missing modality setting for Epic-Kitchens and Something-Something, respectively.

\subsection{Comparison With the State of the Art} \label{sec:comparison-with-sota}
Since KARMMA is inspired by Radevski et al.~\cite{radevski2023multimodal}, which is the current state-of-the-art (SOTA) method in multimodal-to-unimodal distillation for egocentric action recognition, \cref{tab:comparison-with-sota} compares both methods on the Epic-Kitchens dataset. 
For a fair comparison, we report two versions of our model: a unimodal (RGB video only) student (KARMMA$_\text{V}$), and our full multimodal student (KARMMA$_\text{S}$). 

\begin{table*}[ht]
\centering
\small
\caption{\textbf{Comparison with the SOTA multimodal-to-unimodal distillation approach for egocentric action recognition.} 
All methods distill knowledge from a multimodal teacher trained on RGB video (V), optical flow (F), and audio (A). 
``KARMMA$_\text{V}$'' denotes a unimodal version of our multimodal student (``KARMMA$_\text{S}$''). 
Results for Radevski et al. \cite{radevski2023multimodal} are computed using their publicly available student checkpoint. 
\hcolor{} highlights our final student. 
\textbf{Bold} and \underline{underline} values indicate the best and second-best results.}
\label{tab:comparison-with-sota}
\vspace{-0.2em}
\begin{tabularx}{\textwidth}{>{\hspace{-\tabcolsep}}lCCcS[table-format=2.2]S[table-format=2.2]S[table-format=2.2]S[table-format=3]}
\toprule
\textbf{Method} & 
\textbf{\makecell{Student \\ Train. Mods.}} & 
{\makecell{\textbf{Student}\\ \textbf{Infer. Mods.}}} &
{\makecell{\textbf{Flexible Mod.} \\ \textbf{Inference}}} &
{\makecell{\textbf{Teacher Train.} \\ \textbf{Params. (M)}}} & 
{\makecell{\textbf{Student Train.} \\ \textbf{Params. (M)}}}  &
{\makecell{\textbf{Action} \\ \textbf{Acc. (\%)}}} &
{\makecell{\footnotesize{\textbf{GFLOPs}}}} \\
\midrule
Radevski et al. \cite{radevski2023multimodal} & \multirow{2}{*}{V} & \multirow{2}{*}{V} & \xmark & \underline{84.65} & \bfseries 28.21 &  \underline{41.81}  & \bfseries 175\\
KARMMA$_\text{V}$ &  &  & \xmark & \bfseries 11.25 & \underline{29.64} & 40.84 & \underline{176}\\
\midrule
\rowcolor{highlight}
KARMMA$_\text{S}$ & & V & \cmark & \bfseries 11.25 &  65.15 & 38.85  &  \underline{176}\\
\rowcolor{highlight}
KARMMA$_\text{S}$ & \smash{\raisebox{0.15cm}{V+F+A}} & V+F+A & \cmark & \bfseries 11.25 &  65.15 & \underline \bfseries 43.00  &  358\\
\bottomrule
\end{tabularx}
\end{table*}

The top part of~\cref{tab:comparison-with-sota} shows that KARMMA$_\text{V}$ achieves slightly lower accuracy than the student of Radevski et al.~\cite{radevski2023multimodal}. 
However, KARMMA$_\text{V}$ uses a teacher with approximately 7.5 times fewer trainable parameters, as our teacher does not fine-tune its feature extractors.
Moreover, both models are limited to performing inference with RGB video only. 
In robotics, this is problematic as camera occlusions or privacy settings can degrade the RGB stream while other sensors remain available, and training a separate student for each modality combination is impractical on embedded platforms.

The bottom part of~\cref{tab:comparison-with-sota} shows how our KARMMA$_\text{S}$ overcomes this limitation by supporting inference with various modality combinations, unlike the unimodal students. 
Since KARMMA$_\text{S}$ is designed to handle and benefit from multiple modalities, its accuracy on RGB video only is slightly lower than that of the unimodal models, which are optimized specifically for this modality. 
However, when all  modalities are available, KARMMA$_\text{S}$ outperforms the student of Radevski et al. \cite{radevski2023multimodal} by 1.19\,\% (2.9\,\% relative). 
Additionally, when performing inference with a single modality, all models have a similar computational cost, making our multimodal student equally efficient while offering the flexibility to exploit additional sensors when available.
Therefore, KARMMA$_\text{S}$ is applicable across diverse multimodal scenarios, including on-robot deployment, leveraging all available modalities while remaining robust to missing ones.

\subsection{Ablation Studies} \label{sec:ablation-studies}
All ablations are conducted on the Epic-Kitchens-100~\cite{damen2018scaling} dataset with models trained for 50 epochs (\cf, 100 in the main experiments).

\myparagraphnohspace{KARMMA Enhancements.} 
The results in~\cref{tab:analysis-karmma-enhancements} show that applying modality dropout improves the noun, verb and action accuracy of the Baseline. 
Adding our proposed strategy for handling missing modalities further improves verb and action accuracy. 
Knowledge distillation yields the largest gains across all metrics, and extending training to 100 epochs provides an additional boost. 
These results demonstrate that our KARMMA enhancements not only increase the robustness to missing modalities (see~\cref{tab:analysis-of-karmma} and~\cref{fig:impact-missing-modalities-inference}) but also enhance the overall accuracy. 

\begin{table}[t!]
\centering
\small
\caption{\textbf{Analysis of the KARMMA enhancements.} 
``Baseline'' uses the same architecture as our student (see~\cref{sec:teacher-and-student}) but is trained end-to-end with cross-entropy loss and without the KARMMA enhancements (see~\cref{sec:karmma-enhancements}). 
``Longer training'' denotes 100 training epochs. 
\hcolor{} highlights our final student. 
\textbf{Bold} and \underline{underline} values indicate the best and second-best results.
}
\label{tab:analysis-karmma-enhancements}
\vspace{-0.2em}
\begin{tabularx}{\columnwidth}{
>{\hspace{-\tabcolsep}}l@{\hspace{0.8em}}
CCC}
\toprule
\textbf{Enhancement} & 
{\makecell{\textbf{Noun} \\ \textbf{Acc. (\%)}}} & 
{\makecell{\textbf{Verb} \\ \textbf{Acc. (\%)}}} & 
{\makecell{\textbf{Action} \\ \textbf{Acc. (\%)}}} \\
\midrule
Baseline & 50.08 & 64.74 & 38.94 \\
+ Modality dropout & 51.30 & 65.22 & 39.82 \\
+ Missing modality strategy & 50.60 & 65.61 & 39.87 \\
+ Knowledge distillation & \underline{53.05} & \underline{67.24} & \underline{41.68} \\
\rowcolor{highlight}
+ Longer training & \bfseries 54.80 & \bfseries 67.65 & \bfseries 43.00 \\
\bottomrule
\end{tabularx}
\end{table}

\begin{table}[t]
\centering
\small
\caption{\textbf{Impact of knowledge distillation.} 
The ``$\alpha$'' column corresponds to the value used in~\cref{eq:student-loss} that balances the cross-entropy and distillation losses.  
\hcolor{} highlights the value used in our final student. 
\textbf{Bold} and \underline{underline} values indicate the best and second-best results.}
\label{tab:impact-of-KD}
\vspace{-0.2em}
\begin{tabularx}{\columnwidth}{
  C
  S[table-format=2.2]
  S[table-format=2.2]
  S[table-format=2.2]}
\toprule
{$\boldsymbol{\alpha}$} &
{\makecell{\textbf{Noun Acc. (\%)}}} & 
{\makecell{\textbf{Verb Acc. (\%)}}} & 
{\makecell{\textbf{Action Acc. (\%)}}} \\
\midrule
1.0 & 50.60 & 65.61 & 39.87 \\
\rowcolor{highlight}
0.7 & \bfseries 53.05 & \bfseries 67.24 & \bfseries 41.68 \\
0.4 & \underline{52.03} & \underline{66.07} & \underline{40.17} \\
0.1 & 49.98 & 64.51 & 37.82 \\
\bottomrule
\end{tabularx}
\end{table}

\begin{table}[!t]
\centering
\small
\caption{\textbf{Effect of token reduction.} 
``None'' keeps all tokens, while ``$\Theta$-Average'' applies our proposed strategy (see~\cref{sec:token-reduction-strategy}). 
Memory usage refers to the average GPU memory consumed by the fusion block during training. 
The top-1 action recognition accuracy is computed using our teacher, ``KARMMA$_\text{T}$''.
\hcolor{} highlights the configuration used in our final teacher and student. 
\textbf{Bold} and \underline{underline} values indicate the best and second-best results.}
\label{tab:effect-token-reduction}
\vspace{-0.2em}
\begin{tabularx}{\columnwidth}{>{\hspace{-\tabcolsep}}l>{\hspace{-\tabcolsep}\kern-\tabcolsep}C@{\hspace{0.2cm}}c@{\hspace{0.3cm}}S[table-format=2.2]@{\hspace{0.2cm}}S[table-format=2.2]}
\toprule
{\makecell[l]{\textbf{Token} \\ \textbf{Reduction}}} & 
{\makecell{\textbf{\# of Tokens} \\ \textbf{per Mod.}}} &
{\makecell{\textbf{GFLOPs}}} &
{\makecell{\textbf{Mem. Usage} \\ \textbf{of FB (GB)}}} & 
{\makecell{\textbf{Action} \\ \textbf{Acc. (\%)}}} \\
\midrule
None & 785 & 1759 & 14.28 & \bfseries 35.26 \\ 
$\Theta$-Average & 500 & 1725 & 6.35 & 34.76  \\
\rowcolor{highlight}
$\Theta$-Average & \underline{300} & \underline{1707} & \hspace{0.5em}\underline{2.65} & \underline{34.99}  \\
$\Theta$-Average & \bfseries 100 & \bfseries 1693 & \bfseries  0.50 & 32.62  \\
\bottomrule
\end{tabularx}
\end{table}

\myparagraphnohspace{Knowledge Distillation.} 
The balance between teacher supervision (knowledge distillation) and task supervision (cross-entropy) is controlled by the coefficient $\alpha$ in~\cref{eq:student-loss}. 
Therefore, \cref{tab:impact-of-KD} reports the student noun, verb and action accuracy for different $\alpha$ values. 
The results highlight the importance of properly balancing teacher and task supervision: no reliance on the teacher ($\alpha{=}1.0$) fails to leverage the additional guidance it provides, while too much reliance ($\alpha{=}0.1$) leads to overfitting to its outputs and a drop in accuracy. 
The optimal balance is achieved with $\alpha{=}0.7$, yielding an absolute action accuracy gain of 1.81\,\% (4.5\,\% relative) over training without distillation ($\alpha{=}1.0$). 

\myparagraphnohspace{Token Reduction.} 
\cref{tab:effect-token-reduction} shows the impact of token reduction in the fusion block of our teacher. 
Using all tokens yields the highest accuracy, but our parameter-free strategy ($\Theta$\nobreakdash-Average) with $\Theta{=}300$ offers the best accuracy–efficiency trade-off: it reduces memory usage by 11.63\,GB (81.45\,\% relative) and lowers GFLOPs, with only a 0.27\,\% absolute drop in accuracy. 
In more constrained scenarios, a smaller threshold ($\Theta{=}100$) further reduces computational cost while maintaining competitive accuracy.  

\begin{figure}[t]
  \centering
  \includegraphics[width=\linewidth]{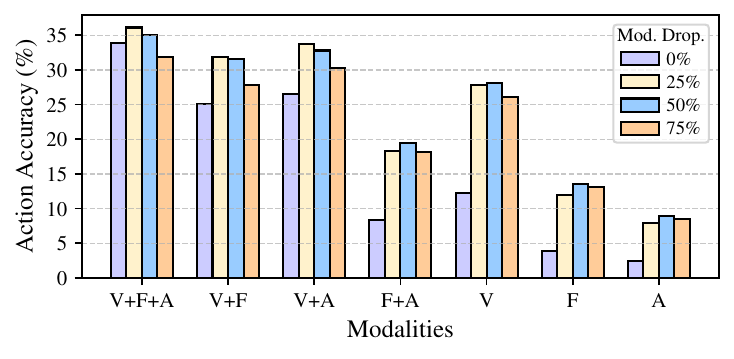}
  \vspace{-2em}
  \caption{\textbf{Impact of modality dropout rate during training.} 
  Each bar shows the top-1 action recognition accuracy of our teacher (see~\cref{sec:teacher-and-student}) trained with a specific modality dropout rate and evaluated across different modality combinations. 
  ``V,'' ``F,'' and ``A'' denote RGB video, optical flow, and audio, respectively. }
  \label{fig:impact-train-mod-drop}
\end{figure}

\myparagraphnohspace{Modality Dropout.} 
During training we apply different modality dropout rates and evaluate their impact at inference. 
\cref{fig:impact-train-mod-drop} reports the resulting action accuracy across various modality combinations, and \cref{tab:impact-train-mod-drop} summarizes these results using three metrics: average action accuracy, average relative drop compared to the full-modality setting (V+F+A), and average rank. 
A 50\,\%  dropout rate achieves the best trade-off, yielding an absolute accuracy gain of 8.15\,\% (50.7\,\% relative) and a 25.42\,\% absolute reduction in relative drop (41.4\,\% relative) compared to no dropout.
These results highlight the importance of selecting an appropriate  dropout rate: too high rates ($\ge 75$\,\%) degrade overall accuracy, while too low rates ($\leq 25$\,\%) fail to build sufficient robustness, resulting in larger accuracy drops when modalities are missing.

\begin{figure}[!t]
  \centering
  \begin{subfigure}{\linewidth}
    \includegraphics[width=1\textwidth]{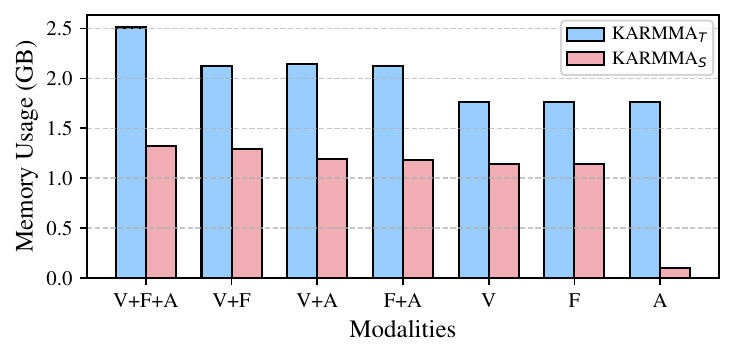}
    \vspace{-1.7em}
  \end{subfigure}
  \begin{subfigure}{\linewidth}
    \includegraphics[width=1\textwidth]{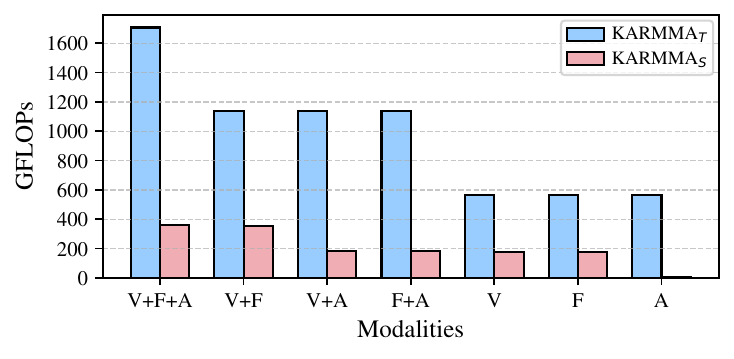}
  \end{subfigure}
  \vspace{-1.9em}
  \caption{\textbf{Resource efficiency of KARMMA student.} 
  ``KARMMA$_\text{T}$'' and ``KARMMA$_\text{S}$'' denote our teacher and student, respectively (see~\cref{sec:teacher-and-student}). 
  The top figure compares their average GPU memory usage for different modality combinations, while the bottom figure compares their GFLOPs, both measured at inference. 
  ``V,'' ``F,'' and ``A'' denote RGB video, optical flow, and audio, respectively.}
   \label{fig:student-efficiency}
\end{figure}

\myparagraphnohspace{Resource Efficiency.} 
\cref{fig:student-efficiency} compares the average memory usage and GFLOPs of the teacher and student at inference time across different modality combinations. 
The results show that, in all cases, our KARMMA student reduces memory consumption by approximately 50\,\% and significantly lowers GFLOPs, resulting in a lightweight model with fast inference. 
The largest savings occur when only audio is used, as the student processes it with an AST-T~\cite{gong2021ast}, which is significantly smaller than the Swin-T~\cite{liu2021swin} used for RGB video and optical flow. 
These savings are especially important for edge and on-robot deployment, where memory budgets are tight and fast inference is desired.

\section{Conclusions}
In this work, we introduced \emph{KARMMA}, a novel multimodal-to-multimodal distillation framework for egocentric action recognition that does not require modality alignment across samples during training or inference. 
The resulting student is lightweight, fast, and benefits from the availability of multiple modalities while remaining robust to missing ones. 
Unlike unimodal approaches, our KARMMA student supports \emph{any} combination of the training modalities without retraining, making it well-suited to real-world multimodal scenarios, including edge and on-robot deployment for applications such as human-robot interaction.

Our framework also reduces training costs by using frozen pre-trained unimodal feature extractors for the teacher, simplifying the integration of newer encoders as they become available. 
Additionally, our simple parameter-free token reduction strategy improves computational efficiency without sacrificing accuracy. 
Overall, KARMMA outperforms the state-of-the-art multimodal-to-unimodal distillation method for egocentric action recognition, achieving higher accuracy while reducing parameter count and GFLOPs. 
Future work will focus on integration into robotic systems and on extending KARMMA to additional modalities to broaden its applicability across platforms and tasks.

\begin{table}[t]
\centering
\small
\caption{\textbf{Summary of the impact of modality dropout rate during training.} 
``Average Accuracy'' refers to the mean action accuracy across all modality combinations. 
``Average Relative Drop'' denotes the mean relative accuracy drop compared to the full-modality setting (V+F+A). 
``Average Rank'' is the mean rank across modality combinations.  
All values are derived from~\cref{fig:impact-train-mod-drop}.
\hcolor{} highlights the modality dropout rate used in our final teacher and student. 
\textbf{Bold} and \underline{underline} values indicate the best and second-best results.}
\label{tab:impact-train-mod-drop}
\vspace{-0.2em}
\begin{tabularx}{\columnwidth}{
>{\hspace{-\tabcolsep}}c@{\hspace{0.9em}}
S[table-format=2.2]
S[table-format=2.2]@{\hspace{1em}}
S[table-format=1.2]
}
\toprule
{\makecell{\textbf{Modality} \\ \textbf{Dropout (\%)}}} &
{\makecell{\textbf{Average} \\ \textbf{Accuracy (\%)}}} & 
{\makecell{\textbf{Average} \\ \textbf{Relative Drop (\%)}}} &
{\makecell{\textbf{Average} \\ \textbf{Rank}}} \\
\midrule
0 & 16.06 & 61.37 & 3.86\\
25 &  \underline{23.97} & 39.24 & \underline{1.86} \\
\rowcolor{highlight}
50 & \bfseries 24.21 & \underline{35.95} & \bfseries 1.43 \\
75 & 22.27 & \bfseries 35.28 & 2.86 \\
\bottomrule
\end{tabularx}
\end{table}

\clearpage

\section*{Acknowledgments}
This work was funded by the Deutsche Forschungsgemeinschaft (DFG, German Research Foundation) under Germany's Excellence Strategy (EXC-3057/1 ``Reasonable Artificial Intelligence'', Project No.\ 533677015).
We gratefully acknowledge support from the hessian.AI Service Center (funded by the Federal Ministry of Research, Technology and Space, BMFTR, grant No.\ 16IS22091) and the hessian.AI Innovation Lab (funded by the Hessian Ministry for Digital Strategy and Innovation, grant No.\ S-DIW04/0013/003).
SSM has been funded by the DFG --~529680848.
MSV, APY, JBC, and JJG further acknowledge support by projects PID2024-158322OB-I00 and PID2021-125209OB-I00, (MCIN/AEI/10.13039/501100011033/ FEDER, UE), project JIUZ2024-IyA-07, DGA 2022-2026 scholarship, and Grant SMT Erasmus+, project 2022-1-ES01-KA131-HED-000065592 funded by Campus Iberus.

\end{document}